%% file: main.tex
\documentclass[preprint,12pt,authoryear]{elsarticle}

\usepackage{amsmath}
\usepackage{amssymb}
\usepackage{bm}
\usepackage{natbib}
\usepackage{graphicx}
\usepackage{url}
\usepackage[ruled,vlined]{algorithm2e}
\usepackage{caption}
\usepackage{subcaption}
\usepackage{multirow}
\usepackage{tabularx}

\usepackage{times}
\usepackage{esvect}
\usepackage{siunitx}
\usepackage{xcolor}
\usepackage{soul}
\usepackage{bookmark}

\makeatletter
\def\ps@pprintTitle{%
	\let\@oddhead\@empty
	\let\@evenhead\@empty
	\let\@oddfoot\@empty
	\let\@evenfoot\@oddfoot
}
\makeatother

\newcommand{\boldx}{\bm{\mathrm{x}}}

\begin{document}
	
\begin{frontmatter}
	
\title{Linear Constraints Learning for Spiking Neurons}
\author{Huy Le Nguyen \corref{cor1}}
\author{Dominique Chu}
\address{CEMS, School of Computing, University of Kent, Canterbury, CT2 7NF, UK}
\cortext[cor1]{Corresponding author. Email address: hln7@kent.ac.uk}

\begin{abstract}%
	We introduce a new supervised learning algorithm based to train spiking neural networks for classification. The algorithm overcomes a limitation of existing multi-spike learning methods: it solves the problem of interference between interacting output spikes during a learning trial. This problem of learning interference causes learning performance in existing approaches to decrease as the number of output spikes increases, and represents an important limitation in existing multi-spike learning approaches. We address learning interference by introducing a novel mechanism to balance the magnitudes of weight adjustments during learning, which in theory allows every spike to simultaneously converge to their desired timings. Our results indicate that our method achieves significantly higher memory capacity and faster convergence compared to existing approaches for multi-spike classification. In the ubiquitous Iris and MNIST datasets, our algorithm achieves competitive predictive performance with state-of-the-art approaches. 
\end{abstract}

\begin{keyword}
	Spiking neural networks, Multi-spike learning, Constraint optimization, Supervised learning,  Classification
\end{keyword}

\end{frontmatter}

\section{Introduction}\label{sec:introduction}
Spiking Neural Networks (SNNs) are a type of neural networks \citep{gk02} that is more biologically plausible compared to the traditional Artificial Neural Networks (ANNs) \citep{prieto16} commonly used in deep learning. In SNNs, a neuron processes and transmits information to others via sequences of discrete events called spikes. This information can typically be encoded in the frequency (rate-coding) or the precise spike timings (temporal coding) of the spike sequences \citep{van2001rate}. In contrast, the output of sigmoidal neurons used in ANNs represents continuous spiking frequencies, and so the operation of such neurons is limited to rate-coding. SNNs with temporal coding have been shown to be more computationally powerful than ANNs \citep{m97}, even on the level of single neurons \citep{rms10}. Furthermore, it has been shown that allowing spiking neurons to generate multiple spikes can increase the diversity, richness, and capacity of information representation in temporal coding schemes \citep{borst1999information,xu2013supervised,pk10}. Therefore, much research into supervised learning for SNNs have focused on developing efficient multi-spike learning algorithms. \citep{pk10,yu2013precise,gardner2016,taherkhani2018supervised,xyyt19,gardner2021supervised}.

One of the main difficulties with multi-spike learning is the problem of learning interference. When learning to generate multiple desired output spikes, updating weights to adjust the timing of one output spike changes the timing of this spike, but inevitably also the timings of other spikes \cite{xu2013supervised,gardner2016}. We will present experimental evidence which suggests that even in the simple case of two output spikes, this learning interference is detrimental to training efficiency (requiring up to ten times more training epochs), or causes non-convergence (weights diverging from their optimal values). A solution to the learning interference problem has not been proposed \cite{taherkhani2020review}, and affected learning algorithms either demonstrate relatively low capacity to memorise (fit) training data \citep{ponulak2005resume, florian2012, mohemmed2012span, yu2013precise, gardner2016}, or achieve high memory capacity but converges slowly \citep{mros14}. For example, the Finite-Precision (FP) algorithm suggested in \cite{mros14} trains SNNs to correctly respond to a large number of inputs, but requires fifty million training epochs for convergence. Another example is the PSD algorithm suggested in \cite{yu2013precise}, which reliably converges in under five hundred epochs, but demonstrates much lower memory capacity compared to the FP algorithm. 

In this paper, we propose a new supervised learning algorithm for training single-layer SNNs to perform classification with multiple spikes, which achieves both high memory capacity and fast convergence. We call this the \textit{Discrete Threshold Assumption} (DTA) algorithm. In precise input-output mapping tasks, our algorithm achieves approximately twice the maximal memory capacity compared to the PSD \citep{yu2013precise} and FILT algorithms \citep{gardner2016}, and converges in several orders of magnitude fewer epochs compared to the FP and HTP algorithms \citep{mros14}. In Tempotron learning tasks \citep{gs06,g16}, our method achieves three times the memory capacity and ten times faster convergence compared to the seminal MST algorithm \citep{g16}. In the practical setting, our algorithm demonstrates competitive accuracy and convergence speed to state-of-the-art methods on Fisher's Iris dataset \citep{fisher1936use} and the popular MNIST dataset \citep{mnist2010}. 

Our method demonstrates that the problem of learning interference in multi-spike learning can be solved for single-layer SNNs, and thus show that learning efficiency does not have to decrease as the number of output spikes increases \citep{xu2013supervised}. Our method achieves this by employing a novel mechanism during learning, which takes the form of a global supervisory signal designed to `balance' the magnitude of weight updates with respect to each output spike time. We formulate this supervisory signal as a linear constraint satisfaction problem, and the basic idea is to compute the temporal correlations between output spike timings. Weight adjustments can then be made such that the effects of learning interference are compensated for, and so already converged spikes are not affected by weight adjustment. In theory, this allows every output spike to converge to their desired timings simultaneously. In practice, the continuous nature of the temporal dimension in the neuron model is problematic, and convergence still requires a number of training iterations. 

The rest of the paper is organised as follows. In Section \ref{sec:neuron}, we describe the neuron model, as well as the definitions of spiking input and output sequences. In Section \ref{sec:learningproblems}, the supervised learning tasks and their notations are introduced. In Section \ref{sec:learninginterference}, the learning interference problem is defined and illustrated by a simple learning scenario. In Section \ref{sec:DTAdescription}, we describe the DTA algorithm in detail. In Section \ref{sec:empiricaltasks}, we evaluate and analyse the memory capacity of our algorithm, compared to various methods in the literature. In Section \ref{sec:practicaltasks}, we evaluate the generalisation performance of our algorithm in practical settings by applying the algorithm to the IRIS and MNIST datasets. Finally, in Section \ref{sec:conclusion} we discuss the DTA algorithm in the context of the wider literature, specifically regarding potential avenues to extend our method to train multi-layer SNNs. 

\section{Results}\label{sec:results}

In this section, we will first introduce the neuron model and the spike-based classification tasks. Then, we will describe the proposed DTA algorithm and how it solves the problem of learning interference for multi-spike problems. 

\subsection{Neuron Model}\label{sec:neuron}

We consider the simplified Spike Response Model (SRM) for its computational simplicity and analytical tractability \citep{gk02}. The model consists of $N \in \mathbb{N^+}$ weighted input channels, an output channel, and an internal state that changes over time. Input spikes to a neuron affect the real-valued internal state $V(t)$ of the neuron (called membrane potential) over time. When $V(t)$ crosses a constant threshold value $\vartheta$ from below, the neuron generates an output spike modelled as the Dirac Delta function $\delta(t)$. An input spike arriving at an input channel at time $t^i \in \mathbb{R^+}$ is modelled as $\delta(t - t^i)$, and an output spike generated by the neuron to its output channel at time $t^o \in \mathbb{R^+}$ is modelled as $\delta(t - t^o)$. 

We denote the set of input spikes to input channel $i$ as $x_i=\{t^i_1, t^i_2, t^i_3, ...\}$. An \textit{input pattern} is defined here as a vector containing specific spike sequences to each input channel, denoted as $\boldx=\{x_1, x_2, ..., x_N\}$. The set of output spike times generated by the neuron is denoted as $o(\vartheta, \boldx, w)=\{t^o_1, t^o_2, t^o_3, ...\}$. Unless stated otherwise, input patterns $\boldx$ are randomly generated and fixed. When the training data is composed of $P \geq 1$ input patterns, the patterns are denoted $\boldx_p$. Each input spike in a spike sequence $x_i$ is generated using a homogenous Poisson point process with rate $\nu_{\mathrm{in}}$ and maximum time value $T \in \mathbb{R^+}$. One of the learning tasks we will consider in this paper requires specifying a desired output spike sequence for each input pattern, the timings of which are generated similarly with rate $\nu_{\mathrm{out}}$.

The momentary membrane potential $V(t)$ of the neuron is defined by the following equations:

\begin{align}
	V(t) &= \sum^N_{i=1}w_i \sum_{t^i \in x_i}\lambda\left(t-t^i\right)-\vartheta\sum_{t^o \in o(\vartheta, \boldx, w)}\gamma(t-t^o)\label{eq:Vs}\\
	\lambda(t-t^i) &=V_{\mathrm {norm}}\left( \exp\left(\frac{-(t-t^i)}{\tau_m}\right)-\exp\left(\frac{-(t-t^i)}{\tau_s}\right)\right)\Theta\left(t-t^i\right)\label{eq:PSPkernel}\\
	\gamma(t-t^o) &= \exp\left( -\frac{t-t^o}{\tau_m}\right)\Theta\left(t-t^o\right)\label{eq:resetkernel}
\end{align}

Here, $\tau_m$ and $\tau_s$ are time constants of the membrane potential and synaptic currents, respectively. The constant $V_{\mathrm {norm}}$ normalises the peak value of function $\lambda(t)$ to 1:

\begin{equation*}
	\begin{split}
		\xi &= \frac{\tau_m}{\tau_s}\\
		V_{\mathrm {norm}} &= \frac{\xi^{\frac{\xi}{\xi-1}}}{\xi-1}
	\end{split}
	\label{Vnorm}
\end{equation*}

And the Heaviside step function:

\begin{equation*}
	\Theta(t) =\begin{cases}
		1, & \text{if $t \geq 0$}\\
		0, & \text{otherwise}
	\end{cases}
\end{equation*}

\begin{figure}[htbp]
	\centering
	\includegraphics[width=0.5\textwidth]{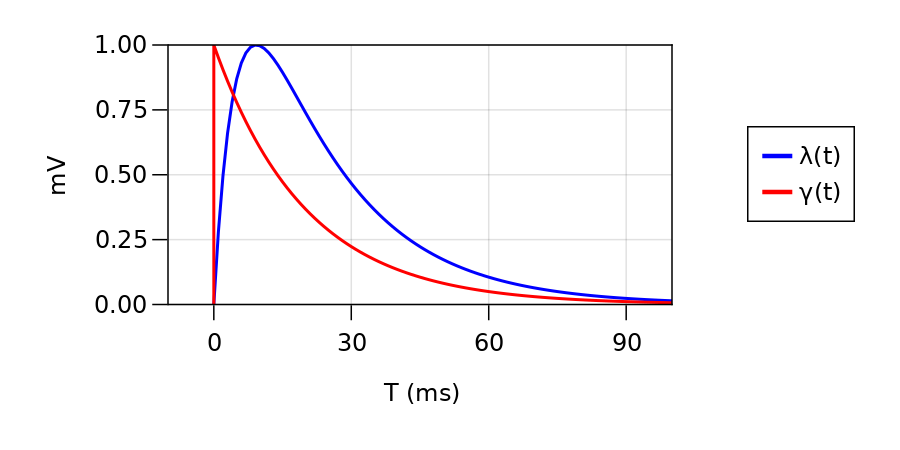}
	\caption{The shapes of the PSP kernel $\lambda$ and the reset kernel $\gamma$ in the neuron model, with $\tau_m=20.0$ and $\tau_s=5.0$.}
	\label{fig:neuronkernels}
\end{figure}

The post-synaptic potential (PSP) function $\lambda(t)$ converts each input spike time into an analog membrane potential contribution that varies over time. The reset function $\gamma(t)$ describes the non-linear reset effect of a previous output spike on $V(t)$. The temporal shapes of $\lambda(t)$ and $\gamma(t)$ are illustrated in Figure \ref{fig:neuronkernels}.

\subsection{Definition of Learning Tasks}\label{sec:learningproblems}

This work will focus on solving multi-spike classification tasks using SNNs with a single trainable layer of spiking neurons. This means the SNN should discriminate fixed input patterns into a number of classes. In traditional ANNs, class predictions are typically decided by the output neuron with the largest (real-valued) output value. Since SNNs do not have real-valued outputs, classification must be decided by some method of output spike encoding, to translate output timings to meaningful class predictions. We will consider two of the most common approaches for classification with multiple spikes. 

In the first method, the supervisory signal provides a specific sequence of desired output spike timings for each input class. This means a trained SNN should respond to an input pattern with spikes at the timings which correspond with the correct class. Classification is decided by the neuron with the most similar output response to the desired sequence, and so requires the calculation of some spike-based similarity measure (we use the Van Rossum metric \citep{rossum2001novel}, see Appendix A). \cite{gutig14} calls this the \emph{Encoding} task, because the network learns to produce mappings between input and output spike sequences. More formally, in the Encoding task the network is given $P$ fixed input patterns $\boldx_p$ and corresponding desired output sequences $y_p=\{t^d_1, t^d_2, t^d_3, ...\}$, and we need to find weights $w$ such that $o(\vartheta, \boldx_p, w)=y_p$ for $p=\{1, 2, ..., P\}$. The Encoding task can be thought of as defining the following constraints on the membrane potential:

\begin{align*}
	\vartheta&=V(t)\text{ at all spike times } t \in y_p \tag{Threshold Equality}\\
	\vartheta&>V(t)\text{ at all other times } t \notin y_p \tag{Threshold Inequality}
\end{align*}

In the second method, the supervisory signal provides a number of desired output spikes for each input class, but no spike timings. As such, the important quantity is the number of output spikes, and these spikes may occur at any time. \cite{gutig14} calls this the \emph{Decoding} task, as the classification can be directly decided by the number of output spikes, without any additional calculations. More formally, in the Decoding task the network is given $P$ fixed input patterns $\boldx_p$ and corresponding labels which define the desired number of output spikes for each pattern $l_p \in \mathbb{N^+}$. Solving this task entails finding weights $w$ such that $|o(\vartheta, \boldx_p, w)| = l_p$ for $p = \{1, 2, ..., P\}$, where $|m|$ denotes the cardinality of the set $m$. 

The Decoding Task can be thought of as an Encoding task with unknown desired spike timings. It follows that in addition to learning to generate the desired spikes, the learning must also find appropriate timings when the spikes should occur. This process has advantages compared to the Encoding task. Specifically, the desired timings in the Encoding task are typically chosen without any specific reason \citep{bohte2002error, booij2005gradient, ghosh2009new}, and so it is difficult to determine whether the desired timings are optimal, such that solutions may exist for one selection of desired timings but not another. In the Decoding task, the freedom to choose output spike timings allows the learning process to identify meaningful input features regardless of when they occur, instead of only those which are temporally close to arbitrarily specified timings. 

\subsection{Learning Interference}\label{sec:learninginterference}
A problem that one encounters when training SNNs to generate multiple output spikes ($l_p > 0$ for Encoding and $|y_p| > 0$ for Decoding) is a phenomenon called `learning interference' \citep{xu2013supervised, taherkhani2020review}. In this section, we will show that learning interference is an important limiting factor for the efficiency of multi-spike learning. We will develop an intuition for learning interference by first considering a first-error weight adjustment approach similar to the FP algorithm suggested by \cite{mros14}. Then, we demonstrate the effects of learning interference in more detail with experimental data, which illustrates the effects of learning interference even in the simple scenario of learning to generate two output spikes. 

For simplicity, we assume the learning task is Encoding. A training algorithm has to adjust neuron weights such that the threshold equality and inequality constraints (as described in the previous section) are satisfied. In continuous time, it is impossible to analytically satisfy an infinite number of threshold inequality constraints, so at each learning iteration we will only consider inequalities at the output spike times that have not converged to desired times. It then follows that there are two ways these constraints can be violated: 

\begin{enumerate}
	\item $\vartheta > V(t^d), \text{ for desired times } t^d \in y_p$
	\item $\vartheta = V(t^o), \text{ for actual times } t^o \notin y_p$
\end{enumerate}

Since $\boldx_p$ and $\vartheta$ are fixed, the only variables that can be adjusted to modify $V(t)$ are the neuron weights. It is evident that we should increase weights contributing to $V(t^d)$ in type (1) violations, to increase $V(t^d)$ to $\vartheta$. Similarly, we should decrease weights contributing to $V(t^o)$ in type (2) violations. These positive and negative weight adjustments can be respectively written as \citep{gs06, mros14}:

\begin{align} \label{eq:naiveupdate}
	\begin{split}
		\Delta w_i^{\mathrm{LTP}}(t^d) &= \eta \sum_{t^i \in x_i}\kappa(t^d-t^i)\\
		\Delta w_i^{\mathrm{LTD}}(t^o) &= -\eta \sum_{t^i \in x_i}\kappa(t^o-t^i)
	\end{split}
\end{align}

Here, $\eta$ is a real-valued learning rate parameter, and $\kappa(t)$ is a temporal learning window which defines the magnitude of $\Delta w_i$ as a function of the input information which is temporally local to the time of constraint violation $t$. In the remainder of this work, we will consider three different $\kappa(t)$ which are commonly used in the literature \citep{yu2013precise,gardner2016}:

\begin{align}
	\kappa_{\mathrm{STDP}}(t) &= \exp\left(\frac{-t}{\tau_m}\right) \Theta(t) \label{eq:kappastdp}\\
	\kappa_{\mathrm{PSP}}(t) &= \lambda(t) \label{eq:kappapsp}\\
	\begin{split} \label{eq:kappafilt}
		\kappa_{\mathrm{FILT}}(t) &= \begin{cases}
			V_{\mathrm{norm}} \left(C_m \exp\left(\frac{-t}{\tau_m}\right) - C_s \exp\left(\frac{-t}{\tau_s}\right)\right), & \text{if $t > 0$}\\
			V_{\mathrm{norm}} \left(C_m-C_s\right) \exp\left(\frac{t}{\tau_m}\right) , & \text{otherwise}
		\end{cases}\\
		C_m &= \frac{\tau_m}{\tau_m+\tau_s}\\
		C_s &= \frac{\tau_s}{\tau_m+\tau_s}
	\end{split}
\end{align}

\begin{figure}[htbp]
	\centering
	\includegraphics[width=0.5\textwidth]{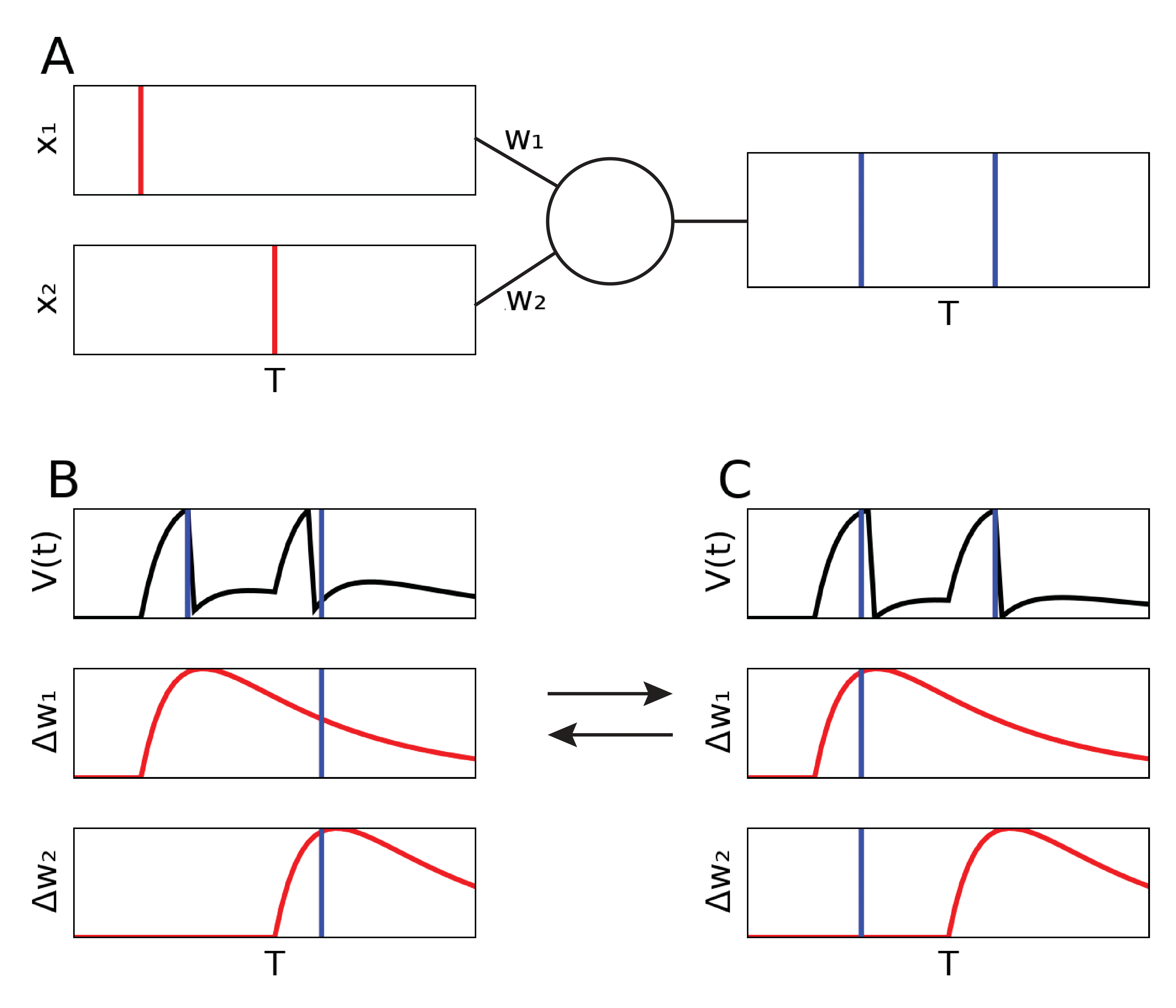}
	\caption{Illustration of learning interference with the first-error weight adjustment approach. \textbf{A}: the Encoding problem with two input spikes (red lines) and two desired output spikes (blue lines). \textbf{B}: (top) First spike is converged, second spike is earlier than desired. Hence, $\Delta w_i^{\mathrm{LTD}}$ is applied which changes both weights; (bottom) intersection between red and blue lines is $\kappa_{\mathrm{PSP}}(t^d)$ for each weight. \textbf{C}: because of the previous weight update, the second spike is converged but the first is now later than desired. Adjusting the first spike with $\Delta w_i^{\mathrm{LTP}}$ then affects the second spike. }
	\label{fig:naiveupdate}
\end{figure}

The first-error approach is to apply either $\Delta w_i^{\mathrm{LTP}}$ or $\Delta w_i^{\mathrm{LTD}}$ as appropriate to only the earliest constraint violation. We now apply this approach to a `toy' problem (shown in Figure \ref{fig:naiveupdate}A), which involves a neuron with two input channels, each receiving one input spike at different timings to each other, and the two desired output spikes should occur shortly after each input spike. Learning interference is defined as the weight adjustment to move one output spike towards its desired timing, having the unintended result of moving other spikes away from their desired timings. This concept is most simply illustrated when one output spike is already converged to the desired time (Figure \ref{fig:naiveupdate}B-C): adjusting the unconverged spike changes the timing of the already converged spike, which then requires re-learning the affected spike. This then affects the originally adjusted spike. In this manner, both weights oscillate around their optimal values, rather than smoothly converging to them. 

\begin{figure}[htbp]
	\centering
	\includegraphics[width=0.5\textwidth]{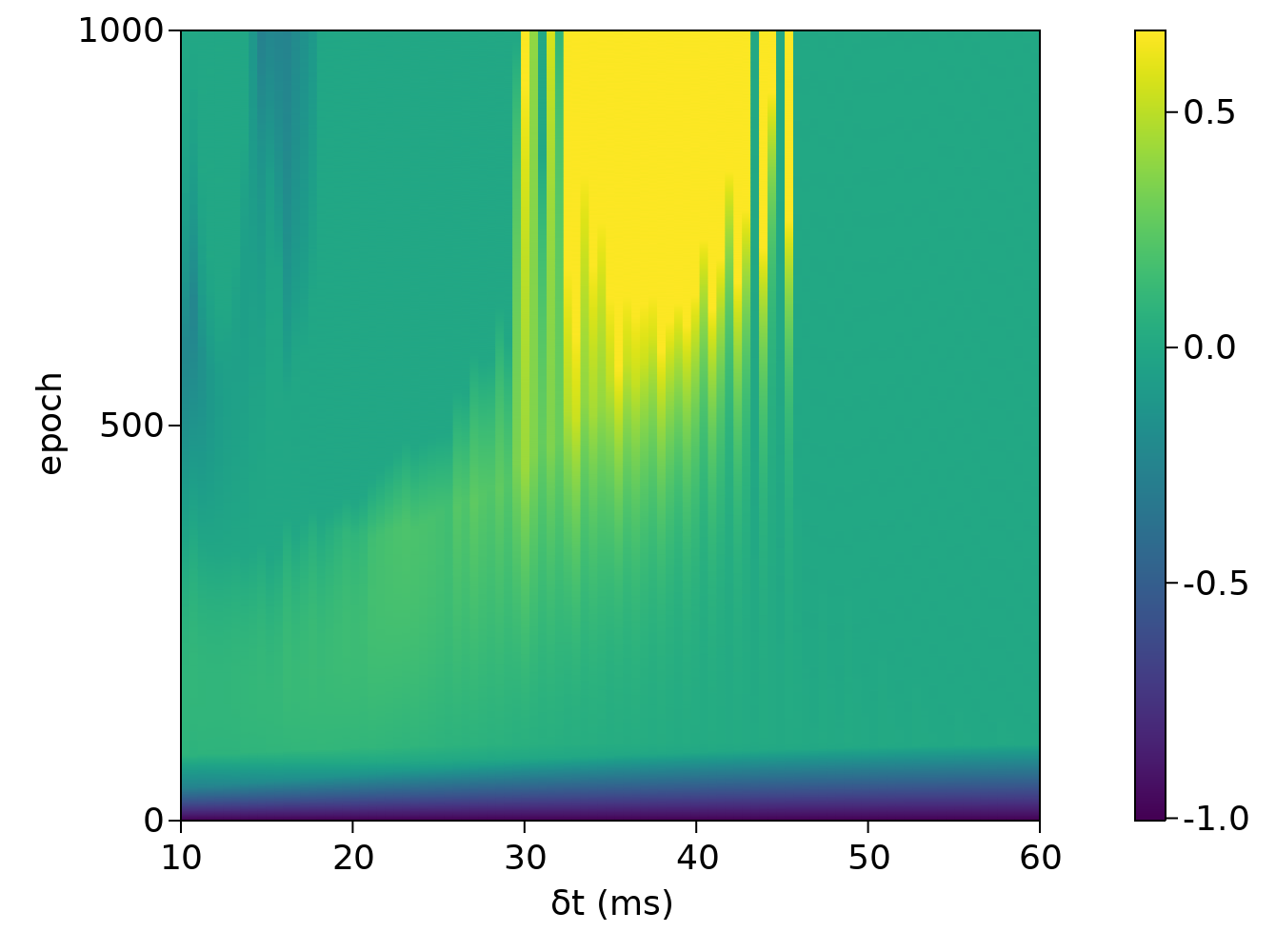}
	\caption{Effects of learning interference on the first neuron weight over the course of training. Vertical axis shows the training epoch. Horizontal axis shows the distance between the two desired output spikes (denoted $\delta t$). Colorbar shows the distance between the weight and its optimal value ($0.0$ is converged). At $\delta t > 50$ no learning interference is visible and the observed weight converges quickly. Otherwise, three effects of learning interference can be observed: at $\delta t < 20$ the weight fluctuates around the optimal value, $20 < \delta t < 30$ eventually converges after many training iterations, and $30 < \delta t < 40$ diverges from the optimal value.}
	\label{fig:limap}
\end{figure} 

The simple Encoding problem above is particularly helpful because if a solution weight vector exists, there can only be one vector in weight space that solves the problem exactly (assuming $t^{i=1} < t^o_1 < t^{i=2} < t^o_2$). For example, the optimal value for the first weight can be obtained by replacing $t$ in Equation \ref{eq:Vs} with the first desired spike time and solving for $w_{i=1}$. By measuring the distance of this weight to its optimal value throughout 1000 epochs of learning, we find that a critical parameter for learning interference is the distance between the two output spikes (Figure \ref{fig:limap}). Here, we observe three different long-term effects of learning interference: (1) weight fluctuating around the optimal value, (2) weight converges, but requires five to ten times more training iterations, and (3) weight diverges from the optimal value. We train the neuron using the PSD method \citep{yu2013precise}, and initial weights are set to zero. This result suggests that even with two output spikes, learning interference can either slow down convergence or prevent it altogether. 

\subsection{DTA Algorithm}\label{sec:DTAdescription}
In this section, we describe the DTA algorithm for training SNNs. We first formulate the algorithm to solve the Encoding task, then describe the (minimal) changes required to solve the Decoding task. During learning, the network is trained for a number of epochs. Each epoch is divided into $P$ training iterations. At each iteration we train the network using one input pattern which is chosen randomly without replacement from the dataset, so an epoch is finished when all input patterns have been presented once. Learning stops if the maximum number of epochs is reached, or the network responds to all input patterns correctly. In the Decoding task, this means responding to each pattern with the correct number of spikes. In the Encoding task, each output spike response to an input should on average be within $T=1$ millisecond from a desired spike (see Appendix A for more detail). 

\subsubsection{Solving the Encoding Task}
We now describe an iteration of the DTA algorithm for solving the Encoding task. The general idea of the algorithm is to convert the weight adjustment calculations into a linear constraint satisfaction problem. Optimising this satisfaction problem involves calculating a ratio of weight adjustments at all desired and actual spike times, with which all threshold constraints are simultaneously satisfied. Thus, in one DTA iteration we will calculate a large number of positive and negative weight adjustments (as in Equation \ref{eq:naiveupdate}) without having to simulate the neuron multiple times, which greatly improves computational efficiency. 

More formally, we begin the learning iteration with weights $w$, and we will calculate weight adjustment $\Delta w$ such that the membrane potential obtained using solution weights $w^* = w + \Delta w$ satisfies all threshold constraints. We replace the parameter $\eta$ in Equation \ref{eq:naiveupdate} with (unknown) variables specific to each output spike time. This means at time $t^d_a \in y_p$ the weight adjustment is multiplied by variable $\eta^d_a$, and at time $t^o_b \in o(\vartheta, \boldx_p, w)$ the adjustment is multiplied by variable $\eta^o_b$. This yields the DTA update equation for a channel $i$:

\begin{equation}\label{eq:batchDTA}
	\Delta w_i = \sum_{a = 1}^{|y_p|} \eta^d_a \sum_{t^i \in x_i}\kappa(t^d_a-t^i) + \sum_{b=1}^{|o(\vartheta, \boldx_p, w)|} \eta^o_b \sum_{t^i \in x_i}\kappa(t^o_b-t^i)
\end{equation}

We calculate appropriate values for $\eta^d_a$ and $\eta^o_b$ by solving a constraint optimisation problem. We first note that in the SRM model it is well established that the reset function can be moved from $V(t)$ to $\vartheta$ without changing the computation of the neuron \citep{gk02}. This means instead of resetting the membrane potential we increase the threshold at each spike time: 

\begin{align}\label{eq:V0s}
	\begin{split}
		V_0(t) &= \sum^N_{i=1}w^*_i \sum_{t^i \in x_i}\lambda\left(t-t^i\right) \\
		\theta(t) &= \vartheta + \vartheta\sum_{t^d \in y} \gamma(t - t^d)
	\end{split}
\end{align}

In this format, the membrane potential can be efficiently calculated using matrix multiplication, which simplifies computation for the optimisation step. The optimisation problem consists of the following linear constraints: 

\begin{align}\label{eq:constraints}
	\begin{split}
		\theta(t)&=V_0(t), t \in y  \\
		\theta(t)&>V_0(t), t \in o(\vartheta, \boldx_p, w) \\
	\end{split}
\end{align}

Additionally, we impose the following domain constraints: 

\begin{align}
	\begin{split}
		lb^d &\leq \eta^d_a \leq ub^d \\
		lb^o &\leq \eta^o_b \leq ub^o
	\end{split}
\end{align}

These domains are necessary to constrain the sizes and directions ($lb^d > 0$ and $ub^o < 0$) of weight adjustments. The size constraint is necessary to minimise membrane potential overshoot, in which the threshold equality constraint is satisfied while $V_0(t)$ is decreasing, rather than increasing. The learning problem now reduces to an optimisation task, which can be solved with standard approaches. In our implementation, we use the Tulip linear optimiser \cite{Tulipjl} which implements the interior-point method \citep{wright2005interior}. Solution weights $w^*$ is used as initial weights for the next iteration, if convergence is not yet reached. 

It is important to also address the non-ideal situation wherein the above optimisation problem returns no feasible solutions. In this scenario, we set all $\eta^d_a$ to a hyper-parameter $\eta=0.001$, and all $\eta^o_b$ to $-\eta$. It follows that the learning iteration where this happens is affected by learning interference, and thus the resulting output spikes may not be at desired timings. However, we find this approach works very well, because the additional threshold inequality constraints allows the problem to become feasible in subsequent learning iterations, and so facilitating an exact solution. This is of course not a problem if a sub-sample of time points are used to impose additional threshold inequalities, however this may significantly increase the dimensionality of the optimisation problem. 

\begin{algorithm}[H]
	\SetAlgoLined
	\KwData{$\bm{\mathrm{X}} = \{\boldx_1, \boldx_2, ..., \boldx_p\}, \bm{\mathrm{Y}} = \{y_1, y_2, ..., y_p\}$}
	initialise neuron with weights $w$\;
	$accuracy \gets 0.0$\;
	\While{epoch $<$ maxEpoch or accuracy $<$ 1.0}{
		shuffle $\bm{\mathrm{X}}$\;
		\ForEach{$\boldx_p$ in $\bm{\mathrm{X}}$}{
			compute $o(\vartheta, \boldx_p, w)$\;
			compute $\Delta w$ (Equations \ref{eq:batchDTA} \& \ref{eq:constraints})\;
			$w^* \gets w + \Delta w$\;
			$w \gets w^*$\;
		}
		$accuracy \gets \frac{\sum_{p=1}^{P} \Theta( vRD^*(T, \Delta t) - vRD(o(\vartheta,\boldx_p, w), y_p))}{P}$ (Appendix A)\;
	}
	\caption{DTA Algorithm for training output neurons in Encoding task}
\end{algorithm}

\subsubsection{Solving the Decoding Task}

\begin{figure}[htbp]
	\centering
	\includegraphics[width=0.6\textwidth]{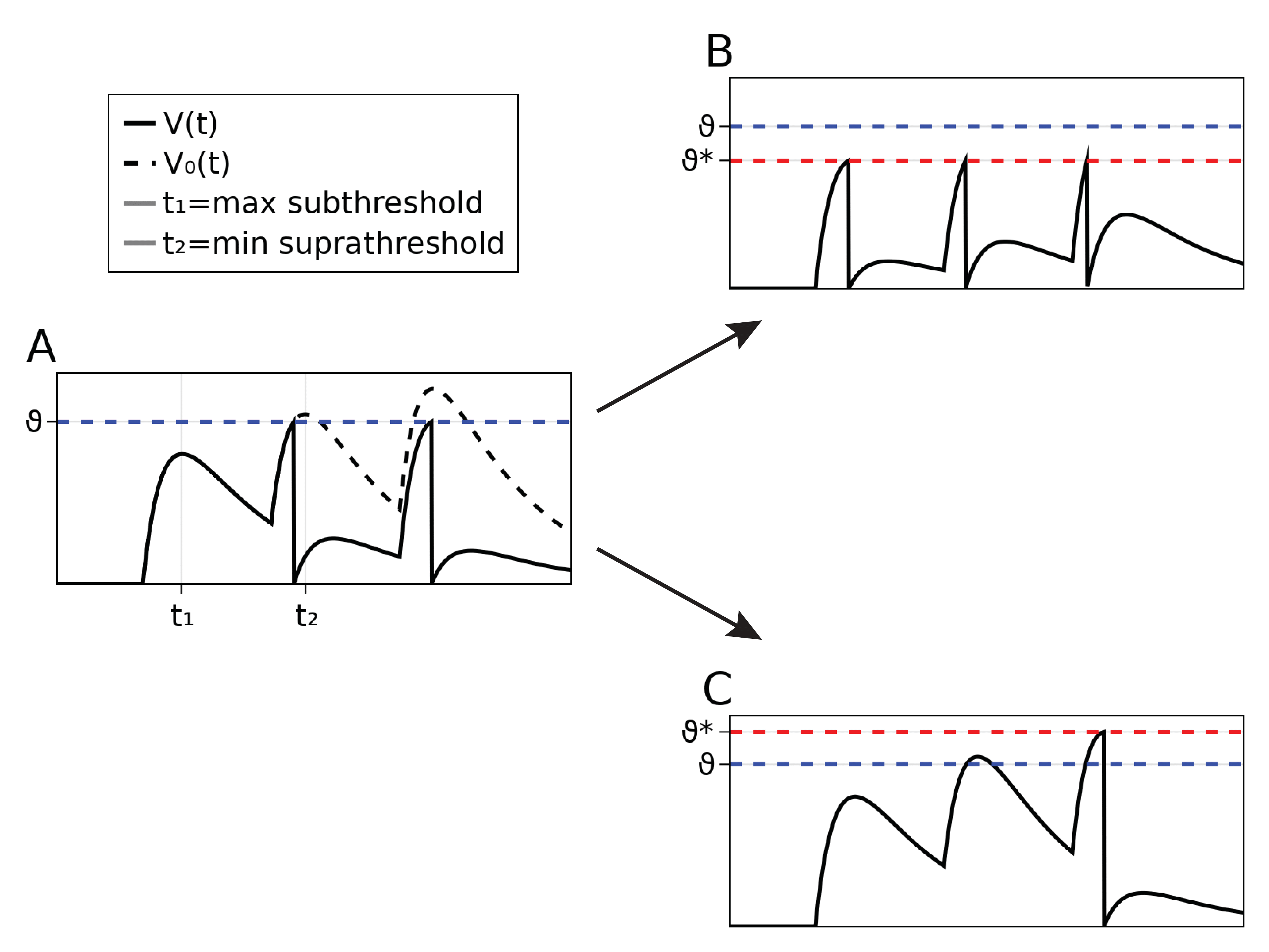}
	\caption{Dynamic threshold procedure for determining desired spike timings in Decoding tasks. \textbf{A}: Membrane potential $V(t)$ (solid line) and membrane potential without reset $V_0(t)$ (dashed line) of a neuron with two output spikes. The appropriate time to generate a new spike is $t_1$, since $V(t_1)$ is closest to $\vartheta$, thus requires the smallest weight adjustment. Similarly, $t_2$ is the appropriate time to remove an existing spike. \textbf{B}: decreasing threshold yields an extra spike at $t_1$. \textbf{C}: increasing threshold removes the spike at $t_2$. We use spike timings in B (3 spikes) and C (1 spike) as desired spike timings, then calculate weight adjustment as with the Encoding task. }
	\label{fig:thetastar}
\end{figure}

We now describe an iteration of the DTA algorithm for solving the Decoding task. The general idea is to first find appropriate desired spike timings for an input pattern, then calculate weight adjustments as in the previous section. The appropriate spike timings are chosen using a `dynamic threshold' procedure, which is inspired by and adapted from Tempotron learning \citep{gs06,g16}. The procedure is as follows: to generate an additional output spike, we lower the spiking threshold to some value $\vartheta^* < \vartheta$, such that the neuron generates one additional spike. Similarly, to elicit one fewer spike, we raise the threshold to $\vartheta^* > \vartheta$. We then simulate the neuron with $\vartheta^*$ to obtain a new set of output spike times $o(\vartheta^*, \boldx_p, w)$, which is used as desired output spike timings. The learning problem now reduces to an Encoding task. This dynamic threshold procedure is illustrated in Figure \ref{fig:thetastar}. 

More formally, we begin the learning iteration by first simulating the neuron to obtain $o(\vartheta, \boldx_p, w)$. If the number of output spikes $|o(\vartheta, \boldx_p, w)|$ is different to the integer label $l_p$, we calculate an appropriate threshold $\vartheta^*$. We obtain $\vartheta^*$ by interval halving in the interval ($0, 10\vartheta$), stopping on the following conditions:

\begin{itemize}
	\item $|o(\vartheta^*, \boldx_p, w)| = |o(\vartheta, \boldx_p, w)| + 1$ if we want more spikes
	\item $|o(\vartheta^*, \boldx_p, w)| = |o(\vartheta, \boldx_p, w)| - 1$ if we want fewer spikes
\end{itemize}

The output spike times $o(\vartheta^*, \boldx_p, w)$ is then chosen as desired spike times for $\boldx_p$. The problem now reduces to an Encoding task, and weight adjustment is calculated as described in the previous section. 

\begin{algorithm}[H]
	\SetAlgoLined
	\KwData{$\bm{\mathrm{X}} = \{\boldx_1, \boldx_2, ..., \boldx_p\}, \bm{\mathrm{Y}} = \{l_1, l_2, ..., l_p\}$}
	initialise neuron with weights $w$\;
	$accuracy \gets 0.0$\;
	\While{$epoch < maxEpoch $ or $ accuracy < 1.0$}{
		shuffle $\bm{\mathrm{X}}$\;
		\ForEach{$\boldx_p \in \bm{\mathrm{X}}$}{
			compute $o(\vartheta, \boldx_p, w)$\;
			\uIf{$|o(\vartheta, \boldx_p, w)| < l_p$}{
				compute $\vartheta^*$ such that $|o(\vartheta^*, \boldx_p, w)| = |o(\vartheta, \boldx_p, w)| + 1$\;
			}
			\uElseIf{$|o(\vartheta, \boldx_p, w)| > l_p$}{
				compute $\vartheta^*$ such that $|o(\vartheta^*, \boldx_p, w)| = |o(\vartheta, \boldx_p, w)| - 1$\;
			}
			$y_p \gets o(\vartheta^*, \boldx_p, w)$\;
			compute $\Delta w$ (Equations \ref{eq:batchDTA} \& \ref{eq:constraints})\;
			$w^* \gets w + \Delta w$\;
			$w \gets w^*$\;
		}
		$accuracy \gets \frac{\sum_{\boldx_p \in \bm{\mathrm{X}}} \delta(|o(\vartheta, \boldx_p, w)| - l_p)}{P}$
	}
	\caption{DTA Algorithm for training output neurons for Decoding task}
\end{algorithm}

\subsection{Memory Capacity}\label{sec:empiricaltasks}

In this section, we evaluate the capacity of the DTA algorithm to memorise (fit) training data. All benchmarks presented here measure the capacity of single spiking neurons. All input patterns for training are randomly generated as described in Section \ref{sec:neuron}. During training, the order of pattern presentation as well as initial weights are controlled across all methods being compared. 

\subsubsection{Capacity in the Encoding Task}\label{sec:encodecapacity}
We first investigate the memory capacity of the DTA algorithm in the Encoding task, compared to existing algorithms in the literature. For random input and output spike sequences, \cite{mros14} has defined memory capacity as the maximal length of inputs $T_\alpha$ that a single spiking neuron can learn. They approximate that this quantity is given by: 

\begin{equation}\label{eq:encodecap}
	T_\alpha \approx \frac{N \sqrt{\tau_m \tau_s}}{\frac{-2 \nu_{\mathrm{out}}\sqrt{\tau_m \tau_s}}{1+\nu_{\mathrm{out}}\sqrt{\tau_m \tau_s}} \mathrm{log}\left(\frac{\nu_{\mathrm{out}}\sqrt{\tau_m \tau_s}}{1+\nu_{\mathrm{out}}\sqrt{\tau_m \tau_s}}\right)} \\
\end{equation}

Here, $\nu_{\mathrm{out}}$ is the output spiking rate, and $N$ is the number of input channels. We will show that the DTA algorithm is able to achieve this maximal memory capacity within 100 epochs of learning. In addition, \cite{mros14} also showed that the memory capacity should be the same for learning multiple short input patterns, and for learning a single long input pattern. For completeness, we will evaluate our algorithm when learning short and long input patterns. In simulations, we compare the performance of our algorithm with the PSD algorithm \citep{yu2013precise}, the FILT algorithm \citep{gardner2016}, and the ReSuMe algorithm \citep{pk10}. These existing algorithms all share the following weight update equation:

\begin{equation*}
	\Delta w_i = \sum_{a = 1}^{|y_p|} \eta \sum_{t^i \in x_i}\kappa(t^d_a-t^i) + \sum_{b=1}^{|o(\vartheta, \boldx_p, w)|} \eta \sum_{t^i \in x_i}\kappa(t^o_b-t^i)
\end{equation*}

In particular, the PSD uses $\kappa_{\mathrm{PSP}}(t)$, FILT uses $\kappa_{\mathrm{FILT}}(t)$, and we implement a version of the ReSuMe algorithm which uses $\kappa_{\mathrm{STDP}}(t)$. The similarity of this update equation with Equation \ref{eq:batchDTA} means that any differences between these algorithms and our DTA method is a direct result of our constraint optimisation formulation. As such, this allows us to demonstrate the performance improvements that is gained from addressing learning interference. 

Our experimental setup is as follows: In the short input scenario, the neuron is trained to memorise short input patterns of length $T=400$. The experiment is repeated over 50 independent trials. In each trial, we begin by training the neuron with only one input pattern. We then subsequently and iteratively increase the number of input patterns, such that the neuron learns two input patterns, then three, and so on. The experiment concludes at the point where half of all experimental trials failed to converge within 500 training epochs, then we calculate $T_\alpha$ as the sum duration of all input patterns. In the long input scenario, the neuron is trained to memorise only a single long input pattern of variable length $T$. In each independent trial, we begin by training the neuron with an input of duration $T=1000$. We then iteratively increase $T$ by increments of 100. The experiment concludes when half of all experimental trials have converged, and $T$ is taken as $T_\alpha$. All algorithms are trained with learning rate hyper-parameter $\eta=0.01$.

\begin{table}[htbp]
	\centering
	\begin{tabular}{
			| c | c | c |
		}
		\hline
		\textbf{Learning Scenario} & \textbf{Method} & $\mathbf{T_\alpha}$ \\
		\hline
		& Theoretical Bound (Equation \ref{eq:encodecap}) & 17200 \\
		\hline
		\multirow{6}{4em}{Multiple Short Patterns} & DTA-$\kappa_{\mathrm{STDP}}$ & 9000 \\
		& DTA-$\kappa_{\mathrm{PSP}}$ & \textbf{18100} \\
		& DTA-$\kappa_{\mathrm{FILT}}$ & 6000 \\
		& ReSuMe-$\kappa_{\mathrm{STDP}}$ & 7500 \\
		& PSP & 5400 \\
		& FILT & 13000 \\
		\hline
		\multirow{6}{4em}{Single Long Pattern} & DTA-$\kappa_{\mathrm{STDP}}$ & 15000 \\
		& DTA-$\kappa_{\mathrm{PSP}}$ & \textbf{17300} \\
		& DTA-$\kappa_{\mathrm{FILT}}$ & 6200 \\
		& ReSuMe-$\kappa_{\mathrm{STDP}}$ & 10900 \\
		& PSP & 7000 \\
		& FILT & 4800 \\
		\hline
	\end{tabular}
	\caption{Maximal memory capacity measurements in the Encoding task, when learning to memorise multiple short input patterns or learning a single long input pattern. The experimental parameters are $N=500, \nu_{\mathrm{in}} = \nu_{\mathrm{out}} = 0.005, \tau_m=20, \tau_s=5$. In both learning scenarios, DTA-$\kappa_{\mathrm{PSP}}$ reaches the theoretical upper bound estimated by \cite{mros14}. We note that standard deviations are not reported in this table, because each data point already represents 50 independent trials.}
	\label{tab:encodecap}
\end{table}

By substituting our parameter values into Equation \ref{eq:encodecap}, the theoretical maximal memory capacity evaluates to $T_\alpha \approx 17200$. Table \ref{tab:encodecap} shows that our algorithm achieves this capacity when $\kappa_{\mathrm{PSP}}$ is used as the kernel function. Notably, our algorithm achieves this desired performance in both learning scenarios, whereas the ReSuMe, PSD, and FILT algorithms all achieve low capacity ($T_\alpha < 10000$) with at least one scenario. To the best of our knowledge, the only other learning algorithms which are capable of reaching this memory capacity are the FP and HTP algorithms \citep{mros14}, and they require fifty million epochs for convergence in their benchmarks.

\begin{figure}[htbp]
	\centering
	\includegraphics[width=0.7\textwidth]{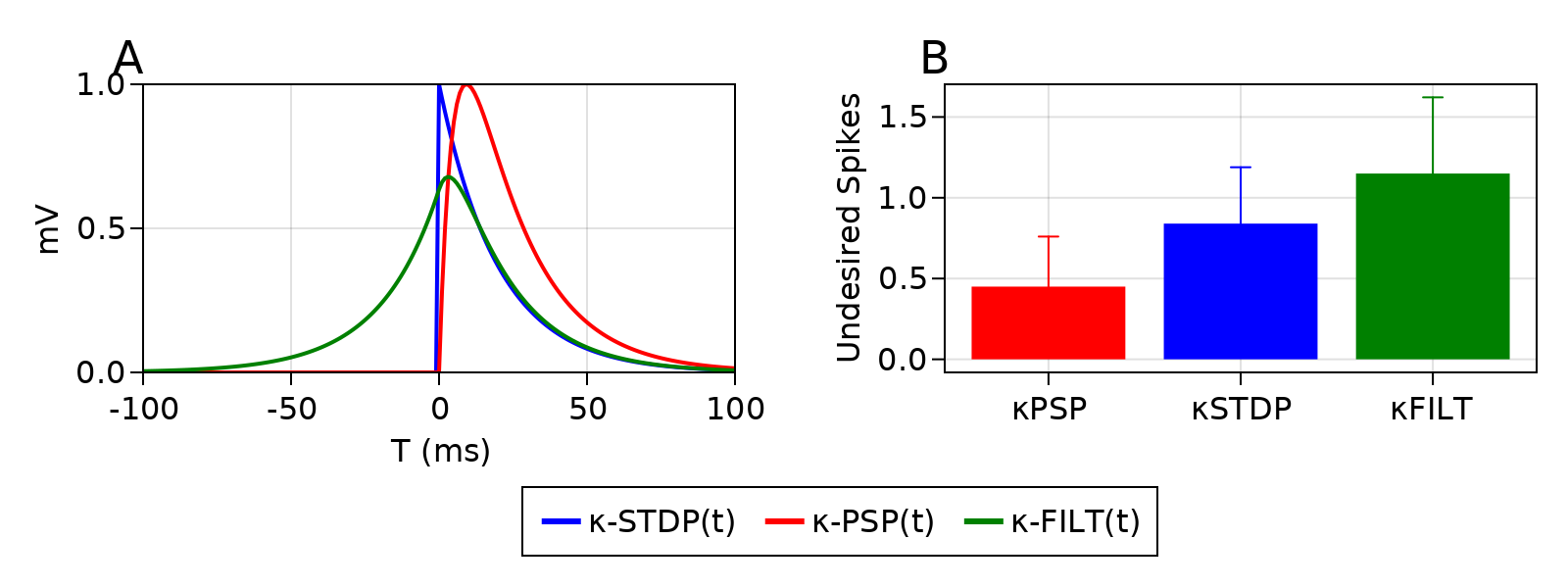}
	\caption{\textbf{A}: The shapes of the kernel functions $\kappa(t)$ used in this work. \textbf{B}: Average number of additional threshold inequality violations (output spikes not at desired timings) after a random number of learning iterations (between 1 and 10) of learning a single input pattern with five desired output spikes, repeated over 5000 independent trials. Error bars are standard deviations.}
	\label{fig:learningkernels}
\end{figure} 

Regarding the reduced performance of our algorithm when using $\kappa_{\mathrm{STDP}}$ and $\kappa_{\mathrm{FILT}}$ as the learning window, we find that these functions cause much more frequent occurrences of threshold inequality violations during learning. This means that after a learning iteration, we observe undesired output spikes which do not occur within one millisecond of a desired spike timing (data shown in Figure \ref{fig:learningkernels}B). There must then be a property of the  $\kappa_{\mathrm{STDP}}$ and $\kappa_{\mathrm{FILT}}$ kernels which results in this undersired behaviour. The most evident characteristic which seperates these kernels from $\kappa_{\mathrm{PSP}}$ is their \textit{anti-causal} property. With $\kappa_{\mathrm{PSP}}$ weights are always adjusted proportionately to their membrane potential contribution; however, with $\kappa_{\mathrm{STDP}}$ and $\kappa_{\mathrm{FILT}}$ weights which do not contribute at all at a desired spike time may be given disproportionately large adjustment. We suspect this causes additional threshold inequality violations at temporally uncorrelated timings. 

\subsubsection{Capacity in the Decoding Task}
We now investigate the memory capacity of the DTA algorithm in the Decoding task, compared to existing algorithms in the literature. Here, we define the memory capacity as the number of different input patterns that a neuron can learn, denoted as $P\alpha$. In this task, the neuron has to discriminate input patterns into five classes, labelled 1 to 5. The experiment is repeated over 50 independent trials. In each trial, each class initially contains only one input pattern. We then evenly and iteratively increase the number of input patterns, such that each class has two patterns, then three, and so on. We stop adding input patterns when half of all trials have failed to converge within 100 epochs, and the experiment concludes at this point. All input patterns are generated with $N=500$ channels, duration $T=50$, and spiking rate $\nu_{\mathrm{in}}=0.005$. 

We compare our algorithm with the MST \citep{g16} and the EMLC \citep{ly20} algorithms, which are both state-of-the-art methods for Decoding. These algorithms are trained with learning rate $\eta=0.01$, with no momentum acceleration. Simulation results are shown in Table \ref{tab:capacitydecode}. Overall, our algorithm achieves approximately twice the maximal memory capacity when compared to the EMLC algorithm, and three times compared to the MST algorithm. 

\begin{table}[htbp]
	\centering
	\caption{Comparison of memory capacity in the Decoding task, when each method is limited to 100 training epochs. Each data point is measured as the point where 50\% of all trials failed to converge.}
	\label{tab:capacitydecode}
	\begin{tabular}{|>{\centering\arraybackslash}p{0.4\textwidth}|>{\centering\arraybackslash}p{0.3\textwidth}|}
		\hline
		\textbf{Method} & \textbf{Memory Capacity}  \\ \cline{1-2} 
		MST \citep{g16}  & 110 \\ \cline{1-2}  
		EMLC \citep{ly20}  & 165 \\ \cline{1-2} 
		DTA-$\kappa_{\mathrm{PSP}}$    & \textbf{355} \\ \hline
	\end{tabular}
\end{table}

\begin{figure}[htbp]
	\centering
	\includegraphics[width=0.55\textwidth]{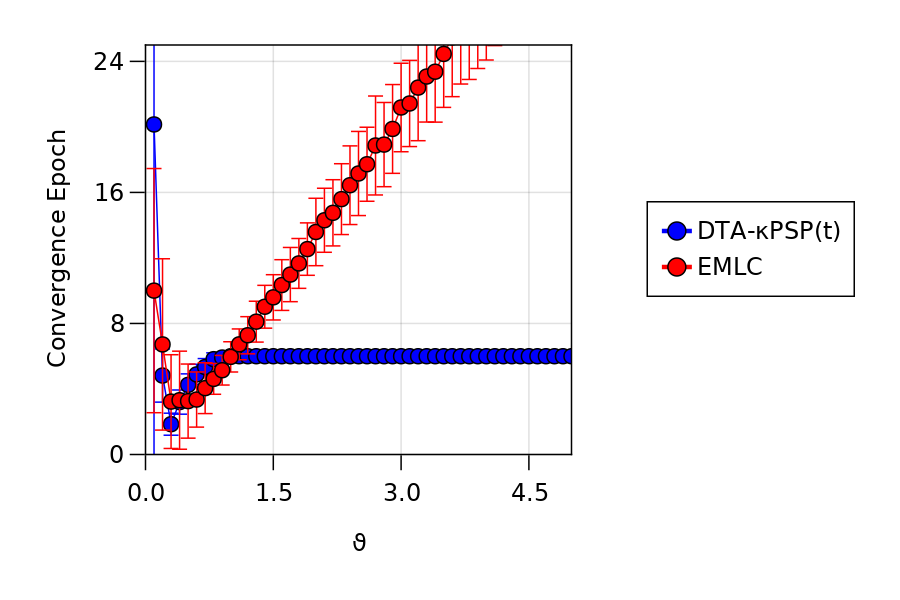}
	\caption{Time complexity scaling of the DTA method compared with the EMLC method \citep*{ly20}, with respect to the spiking threshold $\vartheta$.}
	\label{fig:convergencedecode}
\end{figure}

The improved performance of our method in the Decoding task may appear unusual, because here exact spike timings do not matter and learning interference is only relevant when weight adjustment causes existing spikes to erroneously disappear. To better explain our result, we investigate the difference in complexity scaling of our algorithm compared to the EMLC algorithm. Importantly, we find that our algorithm achieves better complexity scaling with respect to the spiking threshold $\vartheta$. Figure \ref{fig:convergencedecode} shows that our algorithm exhibits constant scaling, whereas the EMLC algorithm scales linearly. This result is intuitive, as while our algorithm changes the number of output spikes by one each iteration, whereas the EMLC algorithm requires a number of iterations to do this (as a linear function of $\vartheta / \eta$). This means that in addition to mitigating effects of learning interference, our treatment of `dynamic' learning rate variables also play an important role in the improved efficiency of our method. 

\subsubsection{Multi-Spike Decoding Capacity of SRM Neuron}
The improved efficiency of our method presents an opportunity to investigate the Decoding capacity of spiking neurons in further detail. The capacity benchmark task presented in the previous section (with a small number of classes) is common in the literature \citep{mtp18, xyyt19, ly20}, however we now introduce a slightly different definition of capacity, denoted as $C_\alpha$. We define $C_\alpha$ as the maximal number of classes that a neuron can learn in the Decoding task, where each class only has a single input pattern. We constrain the class labels to fill the integer range from 1 to $C_\alpha$. This capacity can be interpreted as the ability of the neuron to simultaneously satisfy different output requirements, while the input statistics (input spiking rate) remains constant for all classes. Generally, we are interested in the experimental conditions that maximises $C_\alpha$.

We will measure $C_\alpha$ as follows. In each trial, the neuron initially learns only one class, then we subsequently and iteratively increase the number of classes, such that the neuron learns two, then three classes, and so on. The experiment concludes at the point where 50\% of all independent trials have failed to converge within 100 epochs. Due to the very large computational requirements of this experimental setup, we only train the neuron using the DTA algorithm. 

\begin{figure}[htbp]
	\centering
	\includegraphics[width=0.72\textwidth]{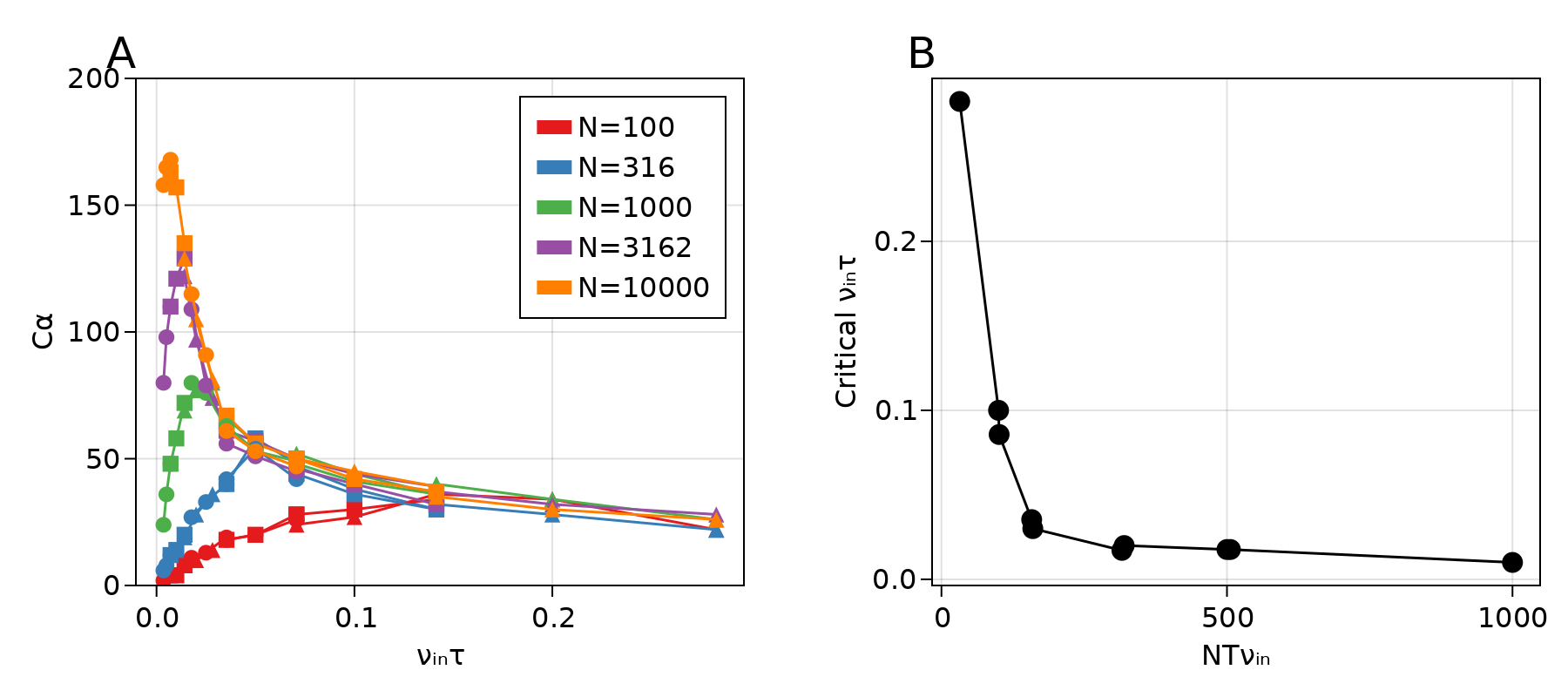}
	\caption{\textbf{A}: Decoding capacity $C_\alpha$, plotted against $\nu_{\mathrm{in}}\tau$ where $\nu_in$ is the input spike rate and $\tau$ is the PSP correlation time. In general, $C_\alpha$ can be expressed as an exponentially decaying function of $\nu_{\mathrm{in}}\tau$. Symbols (circle, square, triangle) respectively represent different series measured with $\tau_m=10, 20, 40$. \textbf{B}: the critical values of $\nu_{\mathrm{in}}\tau$ at which the capacity deviates from the behaviour described in A, plotted against the average number of spikes in an input pattern. Each data point represents 50 independent trials.}
	\label{fig:calphatau}
\end{figure}

Importantly, in the large $N$ limit ($N > 10000$) we find that $C_\alpha$ can be expressed as a function of a parameter $\nu_{\mathrm{in}}\tau$, where $\nu_{\mathrm{in}}$ is the input spiking rate, and $\tau = \sqrt{\tau_m\tau_s}$ is called the PSP correlation time. Specifically, $C_\alpha$ decreases with increasing $\nu_{\mathrm{in}}\tau$ (Figure \ref{fig:calphatau}A). Assuming $\nu_{\mathrm{in}}$ is fixed, $\tau$ should then be small in order to maximise the capacity. This means that the neural dynamics that maximises $C_\alpha$ in the large $N$ limit is such that the membrane potential is able to rise rapidly upon receiving an input spike, and also decays rapidly afterwards. The neuron then operates in a fashion which is reminiscent of coincidence-detection, where each output spike is only generated by coincident inputs during a very small time window. Outside of the large $N$ limit, the capacity quickly deviates from the behaviour described above. More specifically, for $\nu_{\mathrm{in}}\tau$ smaller than a critical value, $C_\alpha$ instead increases with increasing $\nu_{\mathrm{in}}\tau$. We find that the critical value depends only on the average number of spikes in an input pattern (calculated as $NT\nu_{\mathrm{in}}$), as shown in Figure \ref{fig:calphatau}B. 

The above result is consistent with the conclusions published in \cite{mros14} for the Encoding task, and in \cite{rms10} for the binary (two-class) Decoding task. In both of these studies, the memory capacity decreases as $\tau$ increases, assuming that $N$ is large. This then suggests that regardless of the specific learning task, or indeed regardless of whether learning is single-spike or multi-spike, the ability for the membrane potential to rapidly respond to new inputs or forget past inputs is central to the computational capacity of spiking neuron models.

\subsection{Practical Learning Tasks}\label{sec:practicaltasks}

In this section, we evaluate the ability of our algorithm to generalise from training data to unseen examples. We will apply our algorithm to the IRIS dataset \citep{anderson1936species, fisher1936use} and the MNIST dataset \citep{mnist2010}, formulating both experiments as Decoding classification tasks using multiple spikes.

\subsubsection{IRIS Dataset}
Here, we evaluate the generalisation ability of our algorithm on the IRIS dataset, compared to the EMLC \citep{ly20} and MST \citep{g16} algorithms, which are both state-of-the-art Decoding methods. The IRIS dataset consists of 150 samples evenly divided into three classes of different species of the \textit{iris} plant, two classes of which are not linearly separable. Each sample consists of four continuous and real-valued features: petal width, petal length, sepal width, and sepal length. To encode the features into spike times, we use population encoding \citep{bohte2002error} with Gaussian receptive fields of the form $G(v, \mu_j, \sigma)=\exp(-(x-\mu_j)^2/2\sigma^2)$, wherein $v$ is the feature value. Each feature is represented by $M=10$ identically shaped Gaussian functions centered at:

\begin{equation*}
	\mu_j = I^{\mathrm{min}}_j + \left(\frac{2i-3}{2}\right) \left(\frac{I^{\mathrm{max}}_j - I^{\mathrm{min}}_j}{M-2}\right), j \in \{1,2,...,M\}
\end{equation*}

Here, $I^{\mathrm{max}}_j$ and $I^{\mathrm{min}}_j$ represents the maximum and minimum of the $j$-th feature, respectively. The spread of the Gaussian functions are determined by:

\begin{equation*}
	\sigma = \frac{1}{\beta}\left(I^{\mathrm{max}}_j - I^{\mathrm{min}}_j\right)(M-2)
\end{equation*}

Here, $1 \leq \beta \leq 2$ is an adjustment factor, which is set to 1.5 because this value produced the best classification results. As such, each feature value $v$ is converted into $M$ output values $0 \leq G(v, \mu_j, \sigma) \leq 1$, which is then converted into spike times as $t_j = T - G(v, \mu_j, \sigma)T$. In simulation, we choose $T=10$ as this value was determined to provide the best results. 

For learning, we use a SNN with one input layer and one trainable output layer containing three output neurons. Each output neuron is trained to respond to one (target) class of inputs with $s=10$ output spikes, and to remain quiescent for the other two (null classes). A grid-based parameter search showed that generalisation accuracy improves as $s$ increases, but does not improve for $s>10$. Finally, a Winner-Take-All output encoding scheme is applied to the output layer, in which the prediction is represented by the neuron with the most number of spikes. Since our network implements no lateral inhibition between output neurons, the WTA scheme helps improve prediction accuracy on difficult or partially-learnt input patterns, for example when a neuron does not generate precisely $s$ spikes for target patterns or does not remain completely quiescent for null patterns \cite{ly20}. We note also that in any training iteration, if the network prediction is correct \textit{because} of the WTA scheme, weight adjustments are still made to any output neurons with an incorrect number of output spikes (as in the previous example).

\begin{figure}[htbp]
	\centering
	\includegraphics[width=0.5\textwidth]{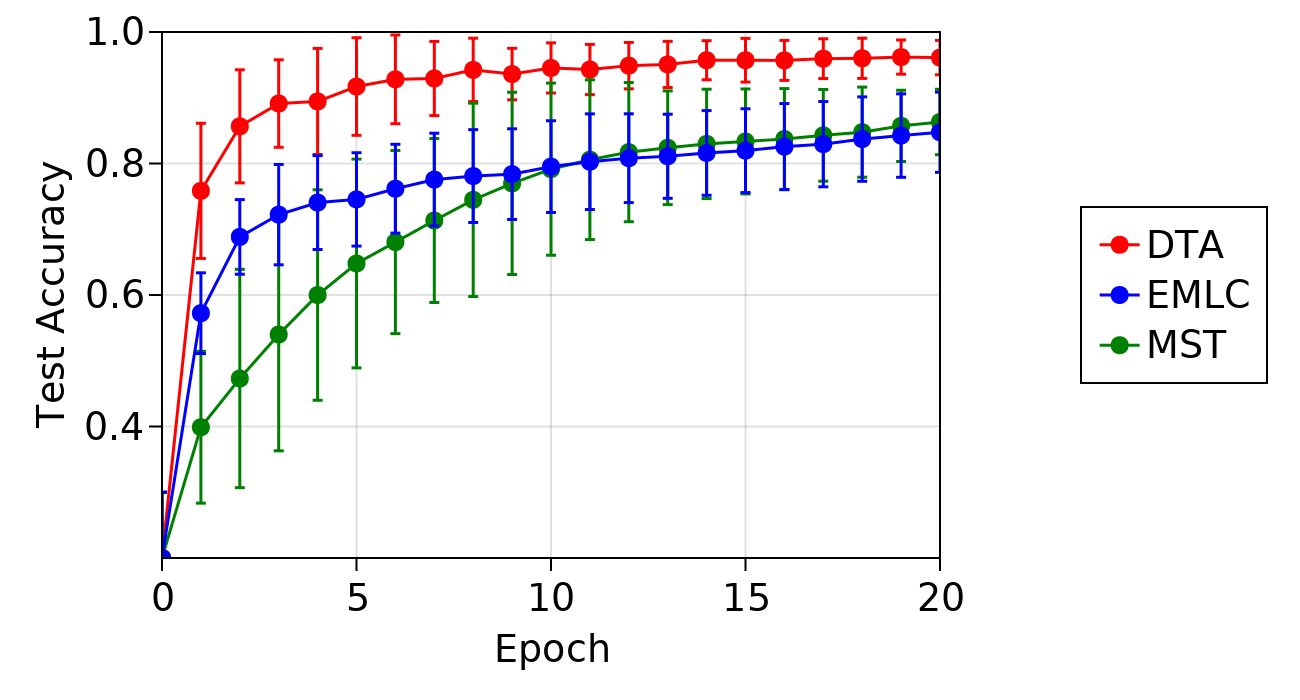}
	\caption{Generalisation accuracy on the IRIS dataset, over 20 epochs of training. Error bars are standard deviations. Each data point represents 50 independent trials.}
	\label{fig:iris}
\end{figure}

The networks are trained for 20 epochs. During each trial, the input patterns are randomly divided into a training set (50\% of patterns) and a testing set (50\% of patterns) for cross-validation. Similarly to the previous section, in each trial we control the order of pattern presentation and the initial weights of the network. Simulation results for 50 independent trials are presented in Table \ref{tab:iris}. In terms of performance, the DTA algorithm achieves approximately 7\% better training accuracy, and approximately 10\% better generalisation accuracy compared to EMLC and MST. In terms of convergence speed, the DTA method also demonstrates an improvement. To achieve the same level of generalisation accuracy that the DTA method achieves after one epoch of training, the EMLC and MST methods both require approximately seven epochs. After the first four training epochs, the DTA method achieves better performance than the final accuracy demonstrated by the EMLC or MST algorithms (Figure \ref{fig:iris}).

\begin{table}[htbp]
	\centering
	\caption{Training and test accuracy of the proposed method on the IRIS flower dataset formulated as a Decoding problem.}
	\label{tab:iris}
	\begin{tabular}{|>{\centering\arraybackslash}p{0.4\textwidth}|>{\centering\arraybackslash}p{0.2\textwidth}|>{\centering\arraybackslash}p{0.2\textwidth}|} 
		\hline
		Method                   		& Train (\%)          & Test (\%)  \\ \hline
		DTA-$\kappa_{\mathrm{PSP}}$		& 96.8 $\pm$ 2.03     & 96.1 $\pm$ 2.61 \\
		EMLC \citep{ly20}               & 89.6 $\pm$ 6.91     & 84.8 $\pm$ 6.10 \\
		MST \cite{g16}                  & 89.4 $\pm$ 2.62  	  & 86.3 $\pm$ 4.98 \\ \hline
	\end{tabular}
\end{table}

The Friedman test was used on the test data, to determine whether or not there is a significant difference between the average ranks of the three algorithms in Table \ref{tab:iris} under the null hypothesis. The calculated $Q$-statistic is $Q=16.8$, which yields a corresponding $p$-value of $p=0.0002$. Thus, we reject the null hypothesis that the three algorithms have no significant differences in generalisation accuracy. The Nemenyi test was used for post-hoc analysis to determine pairwise differences, which reported significant differences between the DTA method, when compared to the EMLC and MST methods (both yielding $p$-values of $0.001$). When comparing the EMLC and MST methods, the statistical difference was not significant ($p=0.425$).

Notably, in this classification task our single-layer network performance is comparable to multi-layer approaches in the literature. In particular, we achieve similar predictive accuracy when compared to \cite{bohte2002error} (reported 96.1\%), \cite{sporea2013supervised} (94\%), and \cite{gardner2021supervised} (95.2\%). The works noted here all utilise a multi-layer architecture with one hidden layer, containing 9-10 hidden neurons. In addition, \cite{bohte2002error} and \cite{gardner2021supervised} uses $M=12$ population input neurons to encode each feature, which is more than our $M=10$. In terms of convergence speed, our method demonstrates an improvement compared to existing methods. For example, \cite{gardner2021supervised, tavanaei2019bp, taherkhani2018supervised} reported high ($> 94\%$) predictive accuracies after 30, 120, and 100 training epochs, respectively. Our setup reaches this performance after the first 8 epochs on average. It is also important to note that while some of the works discussed here also performs classification with multiple output spikes, none of them utilise a Decoding classification scheme, which may also be a factor in the improved accuracy and learning speed which we observe in our results. 

\subsubsection{MNIST Dataset}

In this section, we compare the generalisation accuracy of our algorithm with the EMLC algorithm \citep{ly20}, the MST algorithm \citep{g16}, and traditional gradient-descent using back-propagation. The MNIST dataset consists of images with size $28 \times 28$ pixels, split into ten classes labelled from $0$ to $9$ \citep{mnist2010}. Since the spiking algorithms that we will test are all limited to one trainable layer, we will preprocess the data using a hybrid ANN-SNN framework called CSNN \citep{xqystp18}, and only train the (spiking) output layer. In this way, we only test the capabilities of these algorithms to distinguish input patterns, not feature extraction. 

The CSNN framework combines traditional CNN with a SNN classifier. More formally, the architecture has two layers of rate-coded neurons, and two layers of spiking neurons. Computation through the network can be decomposed into three parts: feature extraction, temporal encoding, and classification. The technical details of this architecture are as follows:

\begin{figure}[htbp]
	\centering
	\includegraphics[width=0.55\textwidth]{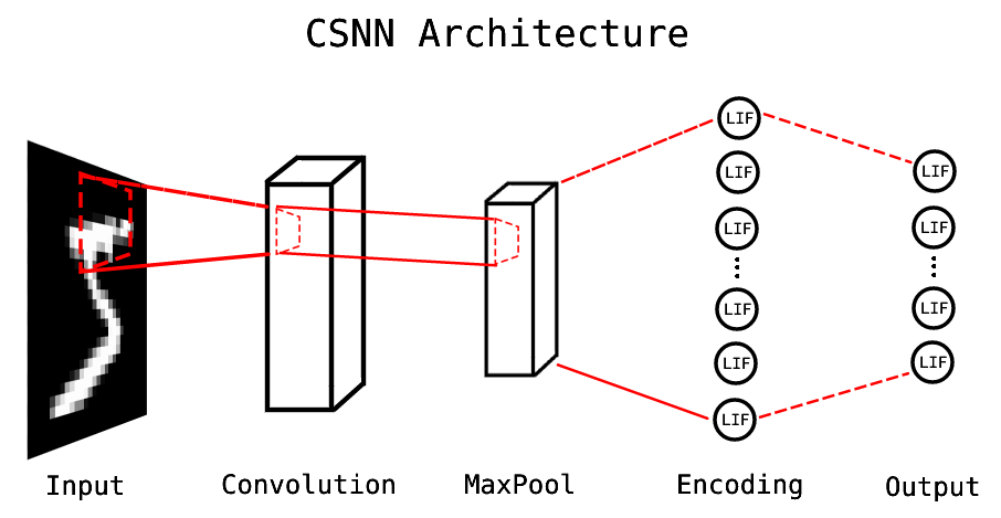}
	\caption{Overview of the CSNN architecture \citep{xqystp18}. The Convolution and MaxPool layers are composed of rate-coded neurons, while the Encoding and Output layers are composed of spiking neurons. In our setup, the Encoding layer has 864 neurons and the Output layer has 10 neurons.}
	\label{fig:CSNNArchitecture}
\end{figure}

First, we train a traditional rate-coded CNN, which provides feature extraction capabilities. The CNN only has three layers: a convolutional layer (6C5), a max-pooling layer (2P2), and a fully connected output layer with 10 neurons. We train this CNN using traditional back-propagation with cross-entropy error function for 30 epochs, then the CNN parameters are fixed and the output layer discarded. The resulting partial-CNN model (the convolutional and pooling layer) performs extraction of invariant local feature maps from the input image.

The feature maps produced by the above partial-CNN must be converted into a spike-based encoding. To this end, the real-valued activations of the pooling layer are linearly mapped to spike times in a time window of length $T=50$. The pooling layer feature maps are flattened to a vector of $864$ activation values. We denote the $i$-th activation value $A_i$ and the corresponding spike time $t^{\mathrm{spike}}_i$. Encoded spike times are calculated as $t^{\mathrm{spike}}_i = T - T \cdot A_i$. These timings are then used as spike times for the encoding layer of LIF neurons. Additionally, any encoding neurons with spike time $t^{\mathrm{spike}}_i=T$ (corresponds to $A_i=0$) do not spike, as their activation is considered too low to induce input spikes.

The encoding layer is fully connected to the output layer, which consists of ten spiking neurons which will be trained. Similarly to the previous section, each neuron is responsible for responding to a `target' class with $s=10$ output spikes, remaining quiescent for all other classes, and a WTA scheme is applied to the output layer. All spiking neurons are initialised with $\tau_m=20.0$ and $\tau_s=5.0$, with initial weights drawn from a Gaussian distribution with mean 0.01 and standard deviation 0.01. 

To evaluate learning performance, we use the full MNIST data of 60000 train and 10000 test images. We train the CSNN described above using the DTA method (CSNN-DTA), the EMLC method (CSNN-EMLC), and the MST method (CSNN-MST). All networks are trained for 20 epochs, with learning rate $\eta=0.001$, without momentum acceleration or learning rate adaptation. As with the previous sections, order of pattern presentation and initial weights are controlled to be the same for all methods in each trial. Simulation results for 10 independent trials comparing these three methods are shown in Table \ref{tab:MNISTData}. 

\begin{figure}[htbp]
	\centering
	\includegraphics[width=0.5\textwidth]{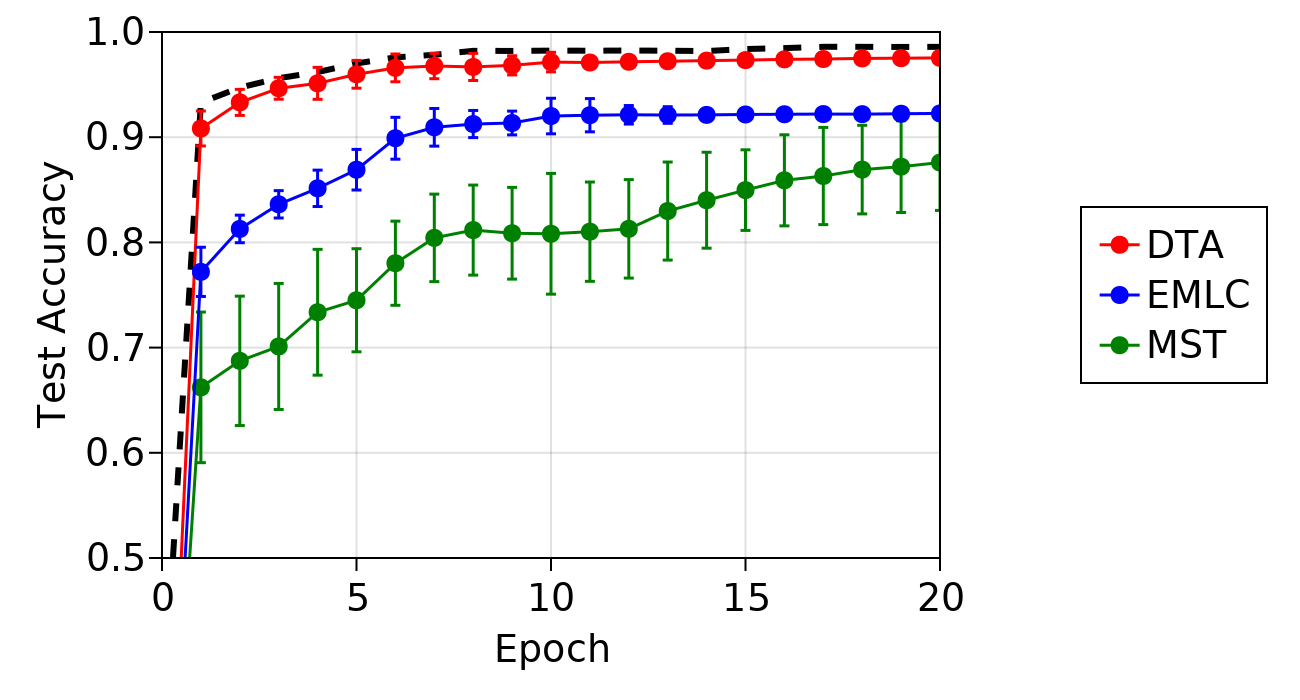}
	\caption{Generalisation accuracy on the MNIST dataset, over 20 epochs of training. Error bars are standard deviations. Each data point represents 10 independent trials. Black dashed lines represents average accuracy of a traditional rate-coded convolutional neural network with the convolution and pooling weights fixed to the same weights which are used in the CSNN.}
	\label{fig:mnist}
\end{figure}

Overall, the DTA method outperforms the EMLC and MST methods in both accuracy and convergence speed. Regarding the generalisation performance, the DTA method achieves approximately 5\% better accuracy compared to EMLC, and approximately 10\% better compared to MST. For good generalisation performance, it is often useful to consider early-stopping in order to prevent over-fitting to training data. If we consider a suitable early-stopping point to be within 1\% of the final accuracies reported in Table \ref{tab:MNISTData}, then the DTA method only requires approximately six training epochs, the EMLC method requires ten epochs, and the MST method requires 18 epochs. This means that the DTA method reaches convergence faster than compared to the EMLC or MST. Similarly, convergence to a good ($> 90\%$) solution is achieved by the DTA method in just 1 epoch, compared to 7 epochs for the EMLC method. The MST method does not achieve this accuracy after 20 epochs of training (Figure \ref{fig:mnist}). 

\begin{table}[h]
	\centering
	\begin{tabular}{|c|c|c|}
		\hline
		Method   &  Train accuracy (\%) & Test accuracy (\%)   \\ \hline
		CSNN-DTA  		       &  $97.57 \pm 0.82$ 	& $97.54 \pm 0.44$ \\ 
		CSNN-EMLC \citep{ly20} &  $94.15 \pm 0.26$ 	& $92.27 \pm 0.26$ \\
		CSNN-MST \citep{g16}   &  $89.98 \pm 4.61$ 	& $86.91 \pm 4.26$ \\ \hline
		
	\end{tabular}
	\caption{Performance comparison of CSNN trained with the MST, EMLC, and DTA methods on the MNIST dataset. Each data point is averaged over 10 independent trials.}
	\label{tab:MNISTData}
\end{table}

The Friedman test was used here to determine whether or not there is a significant difference between the average ranks of the three algorithms in Table \ref{tab:MNISTData} under the null hypothesis. The calculated Friedman statistic is $F_F=47.25$. With three treatments (methods) and 10 independent trials, $F_F$ is distributed according to the $F$-distribution with 3 and 18 degrees of freedom, which for 5\% significant level yields the critical value of 3.16. Since $F_F$ is greater than the critical value, the null hypothesis can be rejected and the differences between the three methods is considered significant. The Nemenyi test was used for post-hoc analysis, which reported significant differences in the predictive performance between the DTA and EMLC methods (with $p$-value of 0.0199), and between the DTA and MST methods (with $p$-value of 0.001). These results indicate that the DTA method demonstrate statistically significant improvement in generalisation performance compared to EMLC and MST. Additionally, we note that even though the CSNN architecture was fundamentally similar, we did not compare our results with those of \cite{xqystp18}, because they did not train output neurons to fire multiple spikes. 

In addition to comparing the DTA method with other spike-based approaches, we also obtained results from a traditional non-spiking CNN (dashed line in Figure \ref{fig:mnist}). This CNN has one convolution and one pooling layer, the weights of these layers are initialised to be the same as those layers of the CSNN, and kept fixed throughout learning. The pooling layer is fully connected to an output layer of 10 rate-coded sigmoidal neurons, and we train this output layer weights with gradient-descent with cross-entropy loss. A small (maximum 0.1) amount of uniform noise is added to the CNN loss during training for regularisation. The order of pattern presentation is controlled to be the same as training the CSNN. This setup allows us to directly compare our spike-based method with the traditional ANN training approach. Overall, the DTA method demonstrates comparable generalisation accuracy as well as convergence speed to that of the CNN. After one epoch of training, CSNN-DTA is approximately 2\% worse than the CNN, and this gradually improves to a final difference of 1\% after 20 epochs. In comparison, EMLC and MST are approximately 6\% and 11\% worse than the CNN in terms of final accuracy. We found no statistically significant difference between DTA and the rate-coded CNN in terms of final generalisation performance ($p=0.507$). These results suggest that our method is competitive with traditional ANN learning, even without any regularisation strategies. 

\section{Discussions \& Further Work}\label{sec:conclusion}
In this paper, we present an efficient learning algorithm to address the problem of learning interference in multi-spike SNN classification problems. We show that training efficiency can improve significantly if interference is addressed. Learning interference has been previously acknowledged in the literature \citep{xu2013supervised,mros14,gardner2016,xie2016efficient,taherkhani2020review} for both single-layer and multi-layer SNNs. It was suggested that `Bigger PSP, Bigger Adjustment' (BPBA) principle can minimise learning interference \citep{xu2013supervised, lqzc19, zhang2021new}. However, we have shown that the PSD algorithm \cite{yu2013precise} is affected by learning interference, even while adhering to the BPBA principle. Another suggested solution is the first-error approach presented by \cite{mros14}, which successfully mitigates the effects of learning interference in Encoding tasks, at the cost of significantly lower learning efficiency. Other methods which are derived from the Widrow-Hoff rule \citep{widrow1990,ponulak2005resume, florian2012, yu2013precise, gardner2016} does not mitigate learning interference in their derivation, but instead they make modifications to empirical benchmark procedures (such as ensuring output spikes are far apart, or constraining input channels to only fire once). 

We note that the linear constraints optimisation step of our algorithm bears similarities to the CONE method \citep{lee2016cone} for analytically calculating optimal solution weights. In particular, the CONE method also utilise threshold constraints to calculate weights. In addition, the CONE method requires additional objective functions which enforce the causality of weights leading to output spikes. In our algorithm, the kernel functions $\kappa(t)$ takes this role, which simplifies the computation to only linear constraints. The flexibility of the `causality' parameters in the CONE method is maintained by simply choosing different $\kappa(t)$ in our method. The CONE method also has several limitations which do not affect our method: it is only applicable to discrete-time neurons, where the set of threshold inequalities are finite. Also, it is a batch method, which requires all the input and output data to be available at the start of training. This requirement is not always feasible, and computation with the CONE algorithm quickly becomes intractable for larger datasets \citep{lee2016cone}. However, we note that with minor modifications to our algorithm, such as appropriate sub-sampling of the input pattern duration, our method can also be formulated as a batch method. This flexibility is an important property for computational analysis, as batch algorithms are useful for quickly searching the solution space of a fixed-size problem. 

The fundamental difficulty in extending the DTA method to multi-layer applications is in the use of the constraint satisfaction formulation. In a multi-layer SNN, the input spike timings arriving at the output layer from a hidden layer is not fixed, which means Equation \ref{eq:V0s} cannot be easily computed. There are two possible approaches to overcome this difficulty. The first approach is to apply a `spike jitter' to output spike timings of hidden layers similar to \cite{xie2016efficient}. This means the timing of each existing spike from a hidden layer neuron is simply shifted in the direction that minimises the difference between desired and actual output timings. Network training could then be carried out in a layer-wise fashion from the input layer to output layer, letting the shifted timings of one layer serve as the input timings for the following layer. The second approach is to consider each input spike $t_i$ in Equation \ref{eq:V0s} as variables in the constraint satisfaction problem. Feasibility of the solutions then only depends on whether appropriate domain constraints can be determined for $t_i$. Out of these two approaches, we consider the former (spike jitter) to be more viable, as in the second approach the constraint satisfaction would no longer be linear, which increases the computational complexity of the problem greatly. 

The central outcome of this work is that learning interference can be resolved by simultaneous weight adjustments in a global and temporally correlated manner. It is now important to consider this intuition from the gradient-descent perspective, as backpropagation remains highly relevant to training multi-layer networks \citep{tavanaei2019deep}. With backpropagation, network weights are adjusted to minimise a differentiable error function $\mathbb{E}$. As an example, we can consider the error function suggested by \citep{xu2013supervised}:

\begin{equation*}
	\mathbb{E} = \frac{1}{2} \sum_{j = 1}^{|y_p|} \left( t^d_j - t^o_j \right)^2 = \frac{1}{2} \sum_{j = 1}^{|y_p|} \mathbb{E}_j
\end{equation*}

In the above equation, $\mathbb{E}$ is formulated as a sum of different components $\mathbb{E}_j$, with each index $j$ corresponding to a desired output spike time $t^d_j$. Derivation of the above equation will result in a learning rule which applies weight adjustments in a local and temporally uncorrelated fashion (as opposed to the above intuition). By this we mean that for some integer $j$, if $\mathbb{E}_j=0$ then this component contributes nothing to the gradient because the $j$-th spike is already converged. In a similar manner to the first-error approach demonstrated in Figure \ref{fig:naiveupdate}, by minimising a different component $\mathbb{E}_{i \neq j} \neq 0$ we may increase $\mathbb{E}_j$. This means that in the process of locally minimising each component, we may observe a net increase in $\mathbb{E}$, rather than the expected net decrease. This example of learning interference can be easily extended to other error functions in the literature such as \citep{lin2017supervised, taherkhani2018supervised, sporea2013supervised}. This means that regardless of derivation, the learning interference problem exists at the level of the construction of the function $\mathbb{E}$. An analogy to the DTA method would now suggest that to maintain convergence of $t^o_j$ in the above example, we should set $\mathbb{E}_j$ to some non-zero value. The exact calculation of this value is still problematic, as this requires calculating the correlation of $t^o_j$ with every other existing and desired timings. Notwithstanding, the significant efficiency improvements of our method in the single-layer SNN application suggests that resolving learning interference is also a worthwhile problem for designing more efficient multi-layer SNNs. 

\clearpage
\bibliographystyle{apalike}
\bibliography{main}

\clearpage
\appendix
\input{appendix}

\end{document}

%% file: appendix.tex
\section{Spike Train Distance Metric}\label{sec:vRD}
In this appendix, we describe the spike train distance metric used in Encoding tasks to quantify the difference between desired and actual output sequences. In this paper, we use the van Rossum distance \citep{rossum2001novel}. This measure is denoted as $vRD(o(\vartheta, \boldx_p, w), y_p)$ for a particular input pattern and output sequence. This measure provides a positive distance between spike sequences, with $vRD(o(\vartheta, \boldx, w), y_p)=0$ meaning the actual and desired sequences are identical (perfect convergence). Here, we describe vRD calculation, and how we choose the necessary parameters. The measure is calculated as:

\begin{align*}
	vRD &= \sqrt{\frac{1}{\tau_q} \int_{-\infty}^{\infty} \left[\sum_{t^d \in y} \kappa_{\mathrm{VD}}(t - t^d) - \sum_{t^o \in o(\vartheta, \boldx, w)} \kappa_{\mathrm{VD}}(t - t^o) \right]^2 dt} \\
	\kappa_{\mathrm{VD}}(t) &= \exp\left(\frac{-t}{\tau_q}\right) \Theta(t)
\end{align*}

Here, the kernel $\kappa_{\mathrm{VD}}(t)$ is used to convert sequences of discrete spikes into continuous functions. $\tau_q$ is a parameter which sets the time scale sensitivity of the metric. Based on analyses presented in \cite{rossum2001novel,satuvuori2018spike}, we set $\tau_q=100$ which provides a good sensitivity to both short-term spike jitter and missing/additional spikes. 

It is practically impossible to achieve perfect convergence, due to limitations of computer precision during learning. We must then choose a parameter to denote `sufficiently good' convergence for simulations. We choose this parameter to be the average distance between each pair of actual and desired output spikes, denoted as $\Delta t$, and set $\Delta t=1$ in this work. This means we consider an output sequence to be converged if each output spike is (on average) within 1 millisecond of their desired timing. This allows us to calculate a minimum distance value, denoted as $vRD^*(T, \Delta t)$. Generally, if $vRD(o(\vartheta, \boldx_p, w), y_p) < vRD^*(T, \Delta t)$ for all values of index $p$, then learning is converged. 

\begin{figure}[htbp]
	\centering
	\includegraphics[width=0.6\textwidth]{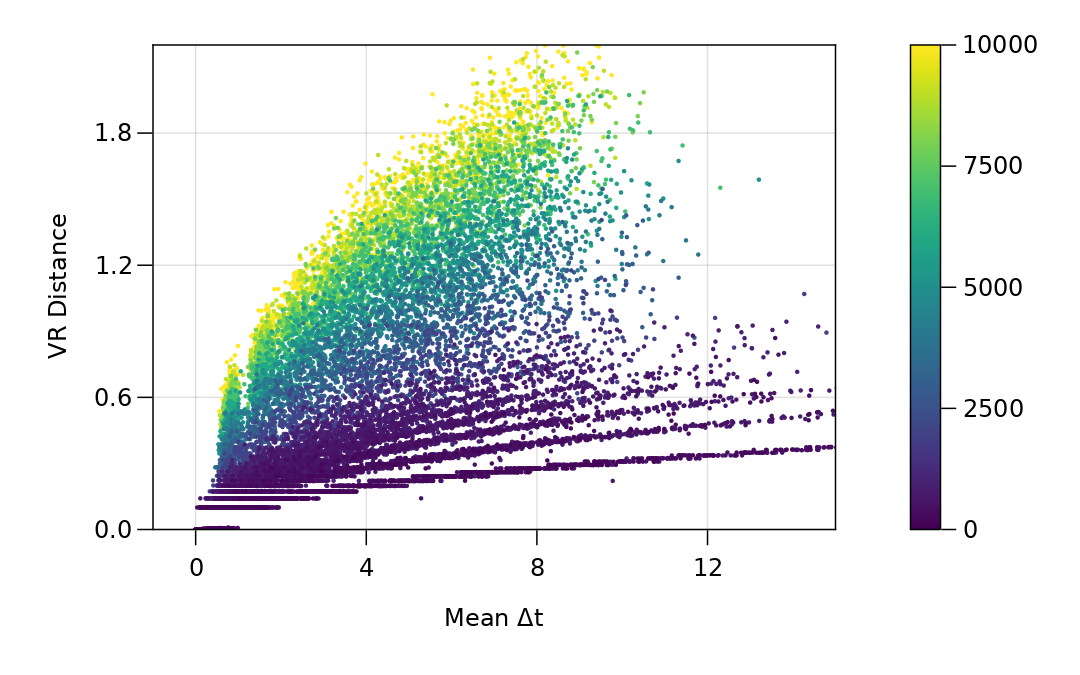}
	\caption{Dependence of van Rossum distance values on the mean spike displacement (x-axis) and pattern duration $T$ (colorbar) with filter time constant $\tau_q=100$.}
	\label{fig:vrd}
\end{figure}

To calculate $vRD^*(T, \Delta t)$, we perform a numerical benchmark to measure the average distance values while varying the distances between actual and desired spike sequences. We first generate 19000 `template' spike sequences with varying duration $100 \leq T \leq 10000$. We then apply a Gaussian spike jitter to each template spike sequence with variance $1 \leq \sigma \leq 10$. The mean difference of template and jittered spike times for each spike sequence are calculated and taken as $\Delta t$. The data is shown in Figure \ref{fig:vrd}. We then fit a Multiple Linear Regression model to the data, where the independent variables are $T$ and $\Delta t$, which yields the following relation:

\begin{equation*}
	vRD^*(T, \Delta t) = 0.08\Delta t + 0.0001T
\end{equation*}

This function is used to calculate the minimum distance value with which we decide convergence. For example, an Encoding task with input patterns of duration $T=1000$ requires the responses to each input pattern to be smaller than $vRD^*(T, \Delta t) = 0.18$.